\documentclass[runningheads]{llncs}
\usepackage{graphicx}
\usepackage{multirow}
\usepackage{array}
\begin{document}
\title{Six-channel Image Representation for Cross-domain Object Detection}
%
%
\author{Tianxiao Zhang\inst{1} \and
Wenchi Ma\inst{1}\and
Guanghui Wang\inst{2*}}
\authorrunning{T. Zhang et al.}
%
\institute{Department of Electrical Engineering and Computer Science, University of Kansas, Lawrence, KS 66045, USA\\
\and
Department of Computer Science, Ryerson University, Toronto, ON, Canada M5B 2K3\\
$*$ Corresponding author. \email{wangcs@ryerson.ca}}

\maketitle              
\begin{abstract}
Most deep learning models are data-driven and the excellent performance is highly dependent on the abundant and diverse datasets. However, it is very hard to obtain and label the datasets of some specific scenes or applications. If we train the detector using the data from one domain, it cannot perform well on the data from another domain due to domain shift, which is one of the big challenges of most object detection models. To address this issue, some image-to-image translation techniques have been employed to generate some fake data of some specific scenes to train the models. With the advent of Generative Adversarial Networks (GANs), we could realize unsupervised image-to-image translation in both directions from a source to a target domain and from the target to the source domain. In this study, we report a new approach to making use of the generated images. We propose to concatenate the original 3-channel images and their corresponding GAN-generated fake images to form 6-channel representations of the dataset, hoping to address the domain shift problem while exploiting the success of available detection models. The idea of augmented data representation may inspire further study on object detection and other applications.


\end{abstract}

\section{Introduction}
Computer vision has progressed rapidly with deep learning techniques and more advanced and accurate models for object detection, image classification, image segmentation, pose estimation, and tracking emerging almost every day \cite{ma2020location,zhang2020real,wu2019unsupervised}. Even though computer vision enters a new era with deep learning, there are still plenty of problems unsolved and domain shift is one of them. Albeit CNN models are dominating the computer vision, their performances often become inferior when testing some unseen data or data from a different domain, which is denoted as domain shift. Since most deep learning models are data-driven and the high-accurate performance is mostly guaranteed by the enormous amount of various data, domain shift often exists when there are not enough labeled specific data but we have to test those kinds of data in the testing set. For instance, although we only detect cars on the roads, training the models on day scenes cannot guarantee an effective detection of cars in the night scenes. We might have to utilize enough datasets from night scenes to train the models, nonetheless, sometimes the datasets from some specific scenes are rare or unlabeled, which makes it even more difficult to mitigate the domain shift effect.

To mitigate the situation where some kinds of training data are none or rare, The image-to-image translation that could translate images from one domain to another is highly desirable. Fortunately, with the advent of Generative Adversarial Networks (GANs) \cite{goodfellow2014generative}, Some researchers aim to generate some fake datasets in specific scenes using GAN models to overcome the lack of data. With some unpaired image-to-image translation GAN models (i.e., CycleGAN \cite{zhu2017unpaired}), it can not only translate images from the source domain to target domain, but also translate images from target domain to source domain, and the entire process does not require any paired images, which make it ideal for real-world applications.

The GAN models for image-to-image translation can generate the corresponding fake images of the target domain from the original images of the source domain in the training dataset, and we can utilize the GAN-generated images to train object detection models and test on images of target domain\cite{arruda2019cross}. Since we expect to solve cross-domain object detection problems, after pre-processing the data and generating the fake images with image-to-image translation models, the generated data has to be fed into the object detection models to train the model and the trained model could demonstrate its effectiveness through testing the data from the target domain. Employing GAN-generated fake images to train the detection models to guarantee the domain of the training data and testing data being the same illustrated the effectiveness of the approach and the detection performance was boosted for the scenario where the training data for the detection models is from one domain while the testing data is in another domain \cite{arruda2019cross}.

\begin{figure}[t]
\centering
\includegraphics[width=0.95\textwidth]{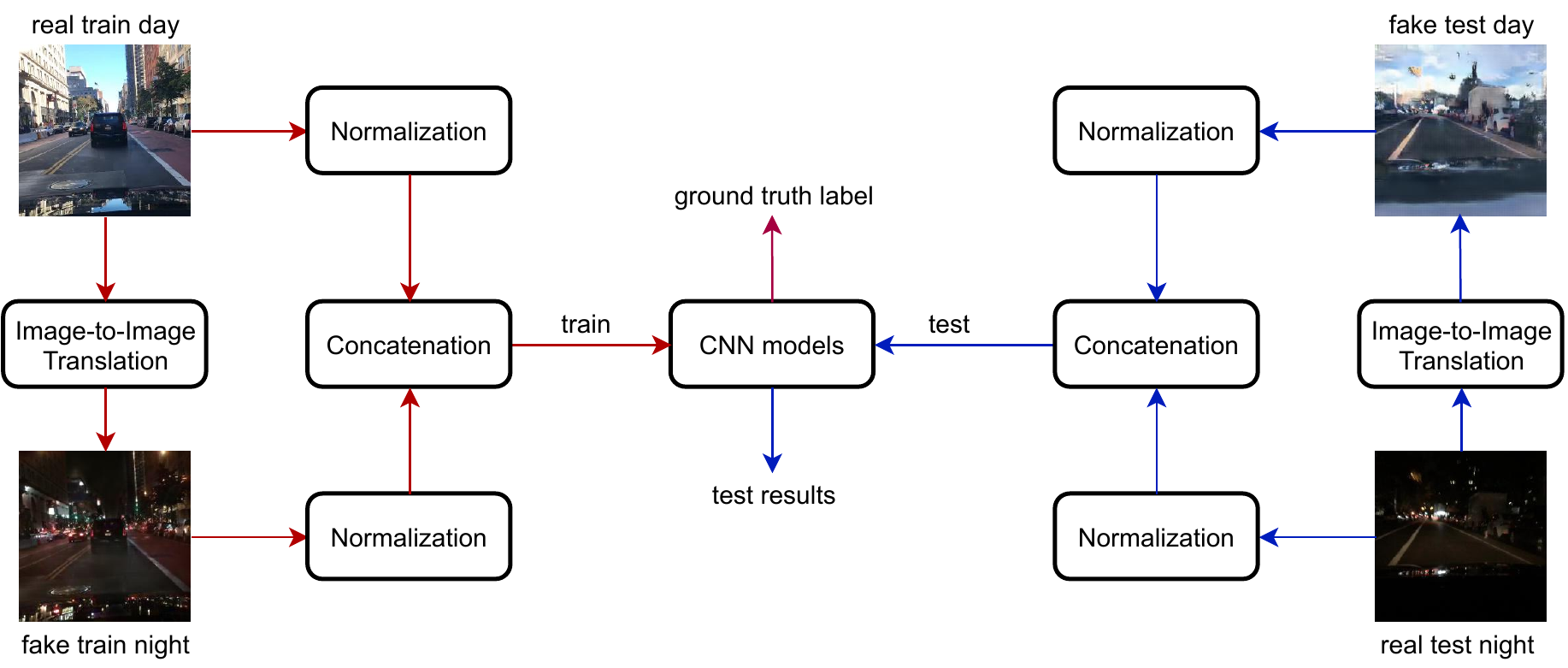}
\caption{The flow chart of the proposed 6-channel image augmentation approach for training and testing CNN-based detection models.}
\label{fig:1}   
\end{figure}

Instead of simply utilizing the fake images to train the model, we propose to solve the problem from a new perspective by concatenating the original images and their corresponding GAN-translated fake images to form new 6-channel representations. For instance, if we only have source domain images but we intend to test our model on unlabeled images in the target domain, what we did was training the image-to-image translation model with source domain data and target domain data. And then we could employ the trained image translation model to generate the corresponding fake images. Since some image-to-image translation models \cite{zhu2017unpaired} could translate images in both directions, we are able to acquire the corresponding fake data for the data from both the source domain and target domain. Thus, both training images and testing images would be augmented into 6-channel representations by concatenating the RGB three channels of the original images with those from the corresponding fake images. Then we can train and test the detection models using available detection models, the only difference is the dimension of the kernel of the CNN models for detection in the first layer becomes 6 instead of 3. The process of training and testing the proposed method is depicted in Fig. \ref{fig:1}.

\section{Related Work}

Image-to-image translation is a popular topic in computer vision \cite{xu2019toward,xu2019stacked}. With the advent of Generative Adversarial Networks \cite{goodfellow2014generative}, it could be mainly categorized as supervised image-to-image translation and unsupervised image-to-image translation \cite{alotaibi2020deep}. The supervised image-to-image translation models such as pix2pix \cite{isola2017image} and BicycleGAN \cite{zhu2017toward}, require image pairs from two or more domains (i.e., the exact same image scenes from day and night), which are extremely expensive and unrealistic to be acquired in the real world. Perhaps the quality of the translated images is sometimes beyond expectations, they are not ideal for real-world applications.

The unsupervised image-to-image translation models can be divided as cycle consistency based models (i.e., CycleGAN \cite{zhu2017unpaired}, DiscoGAN \cite{kim2017learning}, DualGAN \cite{yi2017dualgan}) which introduce cycle consistency losses, autoencoder based models (i.e., UNIT \cite{liu2017unsupervised}) combined with autoencoder \cite{kingma2013auto}, and recent disentangled representation models (i.e., MUNIT \cite{huang2018multimodal}, DIRT \cite{lee2018diverse}). Since the unsupervised image-to-image translation models only require image sets from two or more domains and do not necessitate any paired images which are arduous to collect and annotate, they are often leveraged to generate some fake data in the target domain and applied to other computer vision tasks such as object detection and image classification. Among those unsupervised image-to-image translation models, CycleGAN \cite{zhu2017unpaired} is frequently utilized as the image-mapping model to generate some fake data to be employed in some cross-domain problems \cite{inoue2018cross}\cite{arruda2019cross}.

Object detection addresses the problem that detects the semantic instances on digital images or videos. The fundamental purpose of object detection is to classify the objects shown on the images or videos and simultaneously locate those objects by coordinates \cite{ma2020mdfn}. The applications of object detection are in various fields such as medical image analysis \cite{mo2018efficient}, self-driving car, pose estimation, segmentation, etc. 

From the perspective of stages, the object detectors are categorized into two types: one-stage detectors and two-stage detectors. For two-stage object detectors such as Faster R-CNN \cite{ren2016faster}, MS-CNN \cite{cai2016unified}, R-FCN \cite{dai2016r}, FPN \cite{lin2017feature}, these models are often comprised of a region proposal network as the first stage that selects the candidate anchors which have high probabilities to contain objects and a detection network as the second stage that classify the objects to be contained by these candidates and further do the bounding box regression for these candidates to refine their coordinates and finally output the results. For one-stage object detectors like SSD \cite{liu2016ssd}, YOLOv1-v4 \cite{redmon2016you}\cite{redmon2017yolo9000}\cite{redmon2018yolov3}\cite{bochkovskiy2020yolov4}, RetinaNet \cite{lin2017focal}, these detectors often directly classify and regress the pre-defined anchor boxes instead of choosing some candidates. Thus the two-stage models often outperform the one-stage counterparts while one-stage models frequently have a faster inference rate than two-stage approaches.

Due to the various sizes and shapes of the objects, some models \cite{liu2016ssd}\cite{lin2017feature}\cite{lin2017focal}\cite{zhang2018single} design anchor boxes on different levels of feature maps (the pixels on lower level feature maps have a small receptive field and the pixels on higher-level feature maps have large receptive field) so that the anchors on lower level features are responsible for the relative small objects and the anchors on higher-level features are in charge of detecting relatively large objects. The middle-sized objects are perhaps recognized by the middle-level feature maps.

The aforementioned detection models are anchor-based that we have to design pre-defined anchor boxes for these models. In recent years, some anchor-free models \cite{zhu2019feature}\cite{zhou2019bottom}\cite{duan2019centernet}\cite{tian2019fcos}\cite{law2018cornernet} are attracting great attention for their excellent performance without any pre-defined anchor boxes. Some of them are even dominating the accuracy on COCO benchmark \cite{lin2014microsoft}. Since a large amount of anchors has to be generated for some anchor-based models and most of them are useless because no object is contained in the majority of anchors, anchor-free models might predominate in the designs of object detectors in the future. Recently, the transformer \cite{vaswani2017attention} is applied successfully to object detection \cite{carion2020end}, which is an anchor-free model with attention mechanisms.   

Nonetheless, many problems have not been well solved in this field, especially in cross-domain object detection. Since modern object detectors are based on deep learning techniques and deep learning is data-driven so that the performance of modern object detectors is highly dependent on how many annotated data can be employed as the training set. Cross-domain issues arise when there are not enough labeled training data that have the same domain as the testing data, or the dataset is diverse or composed of various datasets of different domains in both training and testing data.

Domain Adaptive Faster R-CNN\cite{chen2018domain} explores the cross-domain object detection problem based on Faster R-CNN. By utilizing Gradient Reverse Layer (GRL) \cite{ganin2015unsupervised} in an adversarial training manner which is similar to Generative Adversarial Networks (GAN)\cite{goodfellow2014generative}, this paper proposes an image-level adaptation component and an instance-level adaptation component which augment the Faster R-CNN structure to realize domain adaptation. In addition, a consistent regularizer between those two components is to alleviate the effects of the domain shift between different dataset such as KITTI \cite{geiger2013vision}, Cityscapes \cite{cordts2016cityscapes}, Foggy Cityscapes \cite{sakaridis2018semantic}, and SIM10K \cite{johnson2016driving}.

Universal object detection by domain attention \cite{wang2019towards} addresses the universal object detection of various datasets by attention mechanism \cite{vaswani2017attention}. The universal object detection is arduous to realize since the object detection datasets are diverse and there exists a domain shift between them. The paper \cite{hu2018squeeze} proposes a domain adaption module which is comprised of a universal SE adapter bank and a new domain-attention mechanism to realize universal object detection.
\cite{inoue2018cross} deals with cross-domain object detection that instance-level annotations are accessible in the source domain while only image-level labels are available in the target domain. The authors exploit an unpaired image-to-image translation model (CycleGAN \cite{zhu2017unpaired}) to generate fake data in the target domain to fine-tune the trained model which is trained on the data in the source domain. Finally, the model is fine-tuned again on the detected results of the testing data (pseudo-labeling) to make the model even better. 

The study \cite{arruda2019cross} utilizes CycleGAN \cite{zhu2017unpaired} as the image-to-image translation model to translate the images in both directions. The model trained on the fake data in the target domain has better performance than that trained on the original data in the source domain on testing the test data from the target domain. The dataset we employ in this paper is from \cite{arruda2019cross} and we follow exactly the same pre-processing procedure to prepare the dataset. 
In the following, we will discuss our proposal that utilizes concatenated image pairs (real images and corresponding fake images) to train the detection model and compare it to the corresponding approach from \cite{arruda2019cross}.

\section{Proposed Approach}
The framework of our proposed method is depicted in Fig. \ref{fig:1}. In our implementation, we employ CycleGAN for image-to-image translation, which is trained with the data from the source domain (i.e., day images) and the data from the target domain (i.e., night images). First, the fake data (target domain) is generated from the original data (source domain) via the trained image-to-image translation model (i.e., generating the fake night images from the real day images). Then, the real and fake images are normalized and concatenated (i.e., concatenating two 3-channel images to form a 6-channel representation of the image). Finally, the concatenated images are exploited to train the CNN models. During the stage of test, the test data is processed in a similar way as the training data to form concatenated images and sent to the trained CNN model for detection.

\subsection{Image-to-Image Translation}
To realize the cross-domain object detection, we have to collect and annotate the data in the target domain to train the model. While it is difficult to acquire the annotated data in the target domain, image-to-image translation models provide an option to generate fake data in the target domain.

In our experiment, we employed an unpaired image-to-image translation model: CycleGAN \cite{zhu2017unpaired}. CycleGAN is an unsupervised image-to-image translation that only requires images from two different domains (without any image-level or instance-level annotations) to train the model. Furthermore, unpaired translation illustrates that the images from two domains do not need to be paired which is extremely demanding to be obtained. Last but not least, the locations and sizes of the objects on the images should be the same after the image-to-image translation so that any image-level labels and instance-level annotations of the original images can be utilized directly on the translated images. This property is extraordinarily significant since most CNN models are data-driven and the annotations of the images are indispensable to successfully train the supervised CNN models (i.e., most object detection models). Unpaired image-to-image translation models such as CycleGAN \cite{zhu2017unpaired} can translate the images in two directions without changing the key properties of the objects on the images. Thus the annotations such as coordinates and class labels of the objects on the original images can be smoothly exploited in the fake translated images. As manually annotating the images is significantly expensive, by image-to-image translation, the translated images would automatically have the same labels as their original counterparts, which to some extent makes manually annotating images unnecessary.

\begin{figure}[t]
\centering
    \includegraphics[width=.115\textwidth]{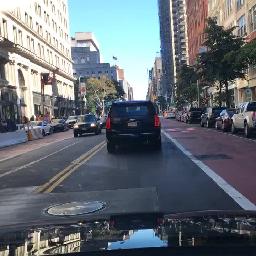}
    \includegraphics[width=.115\textwidth]{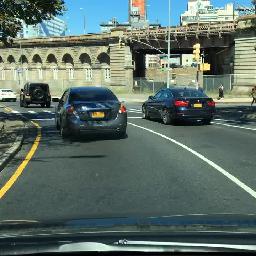}
    \includegraphics[width=.115\textwidth]{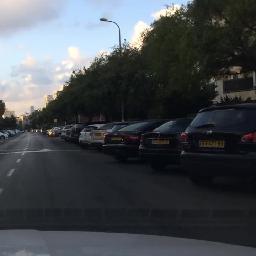}
    \includegraphics[width=.115\textwidth]{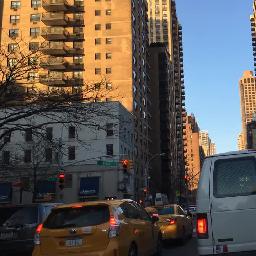}
    \includegraphics[width=.115\textwidth]{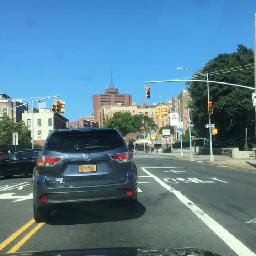}
    \includegraphics[width=.115\textwidth]{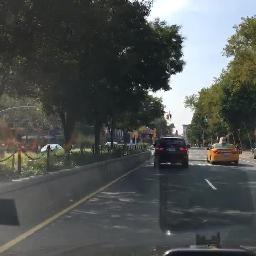}
    \includegraphics[width=.115\textwidth]{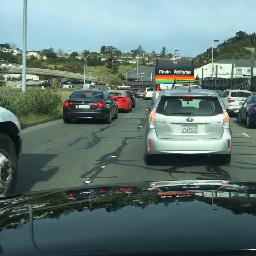}
    \includegraphics[width=.115\textwidth]{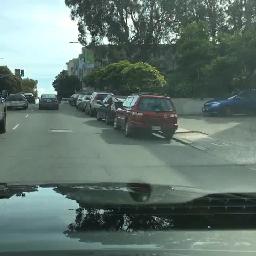}\\
    \includegraphics[width=.115\textwidth]{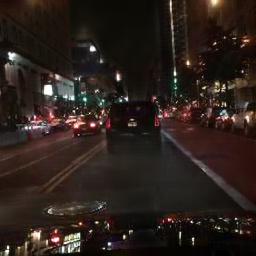}
    \includegraphics[width=.115\textwidth]{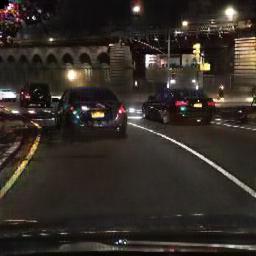}
    \includegraphics[width=.115\textwidth]{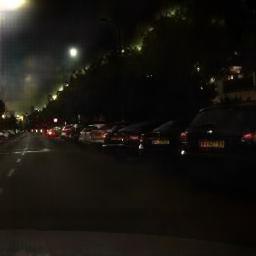}
    \includegraphics[width=.115\textwidth]{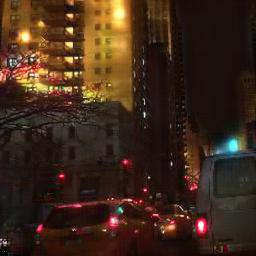}
    \includegraphics[width=.115\textwidth]{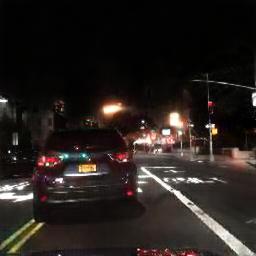}
    \includegraphics[width=.115\textwidth]{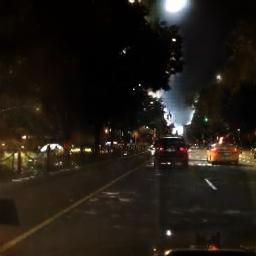}
    \includegraphics[width=.115\textwidth]{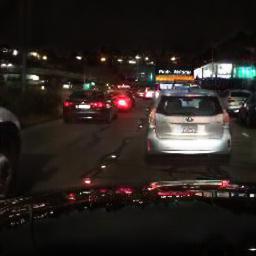}
    \includegraphics[width=.115\textwidth]{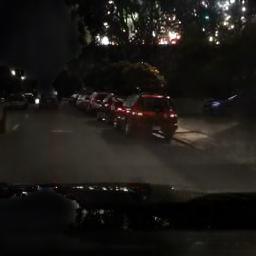}\\
\caption{Several samples of original-day images (1st row) and their corresponding GAN-generated fake-night images (2nd row).}
\label{fig:2}   
\end{figure}

\begin{figure}[t]
\centering
    \includegraphics[width=.115\textwidth]{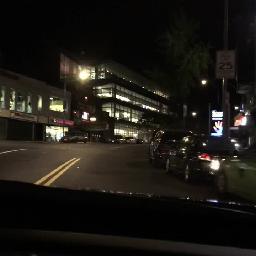}
    \includegraphics[width=.115\textwidth]{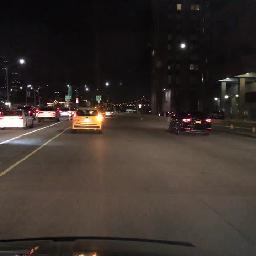}
    \includegraphics[width=.115\textwidth]{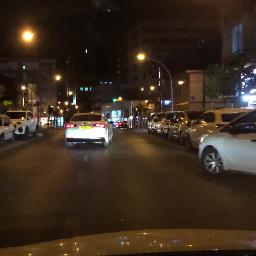}
    \includegraphics[width=.115\textwidth]{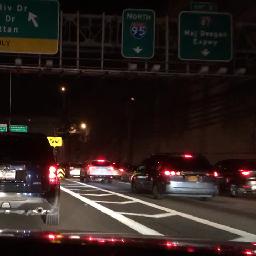}
    \includegraphics[width=.115\textwidth]{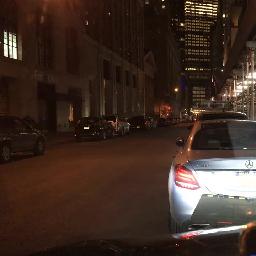}
    \includegraphics[width=.115\textwidth]{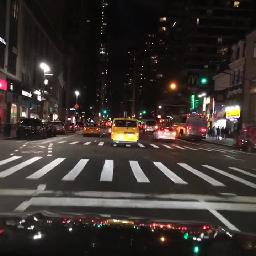}
    \includegraphics[width=.115\textwidth]{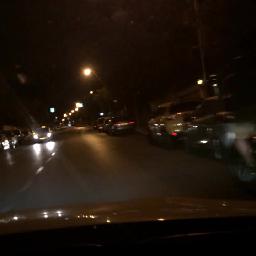}
    \includegraphics[width=.115\textwidth]{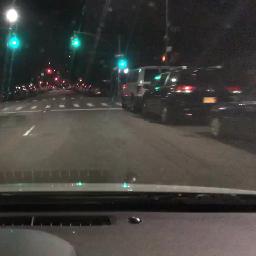}\\
    \includegraphics[width=.115\textwidth]{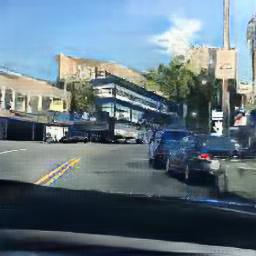}
    \includegraphics[width=.115\textwidth]{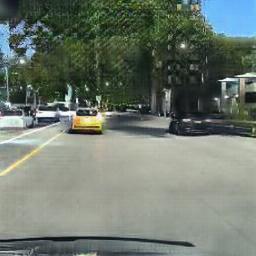}
    \includegraphics[width=.115\textwidth]{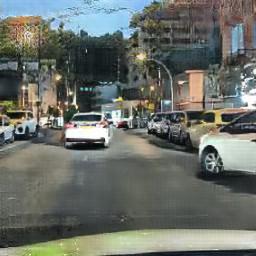}
    \includegraphics[width=.115\textwidth]{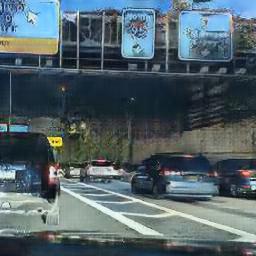}
    \includegraphics[width=.115\textwidth]{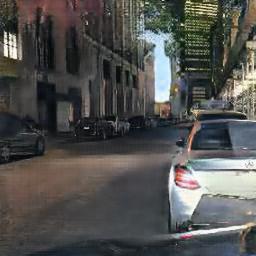}
    \includegraphics[width=.115\textwidth]{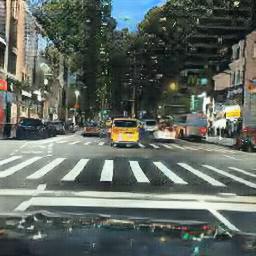}
    \includegraphics[width=.115\textwidth]{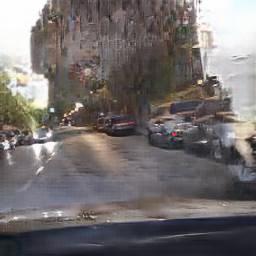}
    \includegraphics[width=.115\textwidth]{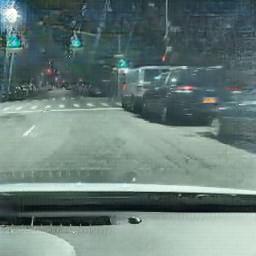}\\
\caption{Several samples of original-night images (1st row) and their corresponding GAN-generated fake-day images (2nd row).}
\label{fig:3}   
\end{figure}

\subsection{CNN Models}
In Fig. \ref{fig:1}, the CNN model can be any CNN-based object detection model, where the dimension of the convolutional kernel in the first layer is changed from 3 to 6. In our implementation, we employ Faster R-CNN \cite{ren2016faster} for detection, and we use ResNet-101 \cite{he2016deep} as the backbone network for the detection model.

Faster R-CNN is a classic two-stage anchor-based object detector that is comprised of Region Proposal Network (RPN) and detection network. Since it is an anchor-based model, we have to design some pre-defined anchor boxes on the feature maps. Typically, 9 anchors with 3 different sizes and 3 different aspect ratios are designed to act as the pre-defined anchor boxes on each location of the feature maps. The objective of RPN is to select some region proposals with a high probability of containing objects from the pre-defined anchors and further refine their coordinates. Each pre-defined anchor would be associated with a score indicating the probability of that anchor box containing an object. Only the anchor boxes with associated scores higher than some threshold can be selected as region proposals and those region proposals are further refined by RPN and later fed into the detection network.

The purpose of the detection network is to receive the region proposals selected and refined by RPN and finally do the classification for each rectangle proposal and bounding box regression to improve the coordinates of the box proposals. Since the region proposals may have various sizes and shapes, more accurately, the number of elements each proposal has might be varying. To guarantee the region proposals are fed into the fully connected layers effectively (the fully connected layer needs the length of input data fixed), the ROI pooling layer is adopted to ensure the size of the input of each proposal to the detection network is fixed. The detection network is simply from Fast R-CNN \cite{girshick2015fast} that is to classify the object which might be contained by each region proposal and simultaneously refine the coordinates of the rectangle boxes. The output of the Faster R-CNN network is the class of the object each proposal might include and the coordinates of the bounding box for each refined proposal. 


\section{Experiments}
In this section, the datasets and the experimental methodology and parameter settings are elaborated. We conducted some of the experiments from \cite{arruda2019cross} for comparison.

\subsection{Datasets}
We employ the same dataset as  \cite{arruda2019cross} in our experiments. The original datasets are from BDD100K \cite{yu2018bdd100k} which is a large-scale diverse dataset for driving scenes. Since the dataset is extremely large and contains high-resolution images and various scenarios on the road and the weather conditions (sunny, rainy, foggy, etc.) \cite{arruda2019cross}, the authors only choose the clear or partly cloudy day and night images to demonstrate the domain shift from day to night \cite{arruda2019cross}. In addition, all selected images are cropped to 256$\times$256 pixels with proper adjustment. There are a total 12,000 images left and processed (6,000 day images and 6,000 night images). After that, the images are randomly sampled and divided into four sets: train-day, train-night, test-day, and test-night, each of the sets contains 3,000 256$\times$256 images. We harness the set of train-day and train-night to train the CycleGAN model and utilized the trained GAN model to generate fake train-night (from train-day), fake train-day (from train-night), fake test-night (from test-day), and fake test-day (from test-night). Now we have a total of 12,000 real images (3,000 for each set) and 12,000 fake images (3,000 for each set). Then we can concatenate the real images and their corresponding fake images to generate 6-channel representations that would be fed into the Faster R-CNN object detector. After choosing and processing the images, the car is the only object on the image to be detected. Some samples of real images and their corresponding GAN-generated fake counterparts are illustrated in Fig. \ref{fig:2} and Fig. \ref{fig:3}.

\subsection{Experimental Evaluations}
Faster R-CNN model is implemented in Python \cite{jjfaster2rcnn} with Pytorch 1.0.0 and CycleGAN is implemented in Python \cite{jyzhu2cyclegan} with PyTorch 1.4.0. All experiments are executed with CUDA 9.1.85 and cuDNN 7 on a single NVIDIA TITAN XP GPU with a memory of 12 GB.

The metric we employed is mean Average Precision (mAP) from PASCAL VOC \cite{everingham2015pascal}, which is the same metric employed in \cite{arruda2019cross}. Since the car is the only object to be detected, the mAP is equivalent to AP in this dataset since mAP calculating the mean AP for all classes.

For CycleGAN, the parameters are default values in \cite{jyzhu2cyclegan}. For Faster R-CNN, similarly to \cite{arruda2019cross}, we utilize pre-trained ResNet-101 \cite{he2016deep} on ImageNet \cite{deng2009imagenet} as our backbone network. We select the initial learning rates from 0.001 to 0.00001 and the experiments are implemented separately for those chosen initial learning rates, but we do not utilize them all for each experiment since our experiments demonstrate that the higher the learning rate we selected from above, the better the results would be. In each 5 epoch, the learning rate decays as 0.1 of the previous learning rate. The training process would be executed 20 to 30 epochs, but the results indicate that the Faster R-CNN model converges relatively early on the dataset. Training every 5 epochs, we record the testing results on test data, but we would report the best one for each experiment. The model parameters are the same for 6-channel experiments and 3-channel experiments, except for 6-channel experiments, the kernel dimension of the first layer of the Faster R-CNN model is 6 instead of 3. And we just concatenate each kernel by itself to create 6-dimension kernels in the first layer of ResNet-101 backbone for 6-channel experiments. While for 3-channel experiments, we simply exploit the original ResNet-101 backbone as our initial training parameters.

\subsection{Experimental Results}
First, we implemented the training and testing of the original 3-channel Faster R-CNN model which is illustrated in Table~\ref{table:1}. The test set is test-night data which is fixed. With different training sets, the detection results on test night are varying.

\setlength{\tabcolsep}{15pt}
\begin{table}[ht]
\centering
\caption{3-channel detection}
\begin{tabular}{ |p{6cm}|p{1cm}|}
 \hline
Train set & mAP\\
 \hline
train-day (3,000 images) & 0.777\\
fake train-night (3,000 images) & 0.893\\
train-night (3,000 images) & 0.933\\
train-day + train-night (6,000 images) & 0.941\\
 \hline
\end{tabular}
\label{table:1}
\end{table}
\setlength{\tabcolsep}{15pt}

From Table~\ref{table:1} we can see that, for testing the test-night set, the model trained on the fake-train night set is much better than that trained on the original train-day set, which corresponds to the results from \cite{arruda2019cross}. These experimental results indicate that if the annotated day images are the only available training data while the test set contains only night images, we could leverage fake night images generated by the image-to-image translation models to train the CNN model. The results are excellent when the model is trained on the train-night set (without domain shift), indicating the domain shift is the most significant influence on the performance of the CNN model in this experiment.

Then we conduct the experiments for our proposed 6-channel Faster R-CNN model which is shown in Table~\ref{table:2}. The test data is comprised of test-night images concatenated with corresponding translated fake test-day images. The training sets in Table~\ref{table:2} have 6 channels. For instance, train-day in the table indicates train-day images concatenated with corresponding fake train-night images, and train-day plus train-night in the table represents train-day images concatenated with corresponding fake train-night images plus train-night images concatenated with corresponding fake train-day images. 

\setlength{\tabcolsep}{15pt}
\begin{table}[ht]
\centering
\caption{6-channel detection}
\begin{tabular}{ |p{9cm}|p{1cm}|}
 \hline
Train set & mAP\\
 \hline
train-day (3,000 6-channel representations) & 0.830\\
train-night (3,000 6-channel representations) & 0.931\\
train-day + train-night (6,000 6-channel representations) & 0.938\\
 \hline
\end{tabular}
\label{table:2}
\end{table}
\setlength{\tabcolsep}{15pt}

From Table~\ref{table:1} and Table~\ref{table:2}, it is noticeable that even though the model trained on train-day images concatenated with fake train-night images (6-channel) has a better result with AP 0.830 than that just training on train-day (3-channel) with AP 0.777, it is worse than the model only trained on fake train-night (3-channel) with AP 0.893.

To demonstrate if the 6-channel approach can improve the detection results in the situation where the training set and testing set do not have domain shift, we also performed the experiment that trains the model on train-night set (3-channel) and tests it on test-night set. From Table~\ref{table:1}, the average precision is 0.933, which is pretty high since there is no domain shift between the training data and testing data. Accordingly, we did the corresponding 6-channel experiment which trains on train-night set concatenated with fake train-day set and tests it on test-night images concatenated with fake test-day images. From Table~\ref{table:2}, the average precision of this 6-channel model is almost the same as its corresponding 3-channel model.

We increase the size of the training data by training the model with the train-day set plus the train-night set and testing it on test-night data. From Table~\ref{table:1} and Table~\ref{table:2}, the result of 6-channel model also performs similar to its 3-channel counterpart. More experimental results are shown in Table~\ref{table:3}, which are from the original 3-channel models. To remove the effect of domain shift, the training set and the testing set do not have domain shift (they are all day images or night images). From Table~\ref{table:3}, it is obvious that the "quality" shift influences the performance of the models. For instance, the model trained on the original train-day (or train-night) set has better performance on the original test-day (or test-night) set than the GAN-generated fake day (or night) images. Similarly, the model which is trained on GAN-generated fake train-day (or fake train-night) set performs better on the GAN-generated fake test-day (or fake test-night) set than the original test-day (or test-night) set.

\setlength{\tabcolsep}{20pt}
\begin{table}[ht]
\centering
\caption{3-channel extra experiments}
\begin{tabular}{ |c|c|c|}
 \hline
Train set & Test set & mAP\\
\hline
\multirow{2}{*}{train-day} & test-day & 0.945\\
& fake test-day & 0.789\\
\hline
\multirow{2}{*}{fake train-day} & fake test-day & 0.914\\
& test-day & 0.903\\
\hline
\multirow{2}{*}{train-night} & test-night & 0.932\\
& fake test-night & 0.859\\
\hline
\multirow{2}{*}{fake train-night} & fake test-night & 0.924\\
& test-night & 0.868\\
\hline
\end{tabular}
\label{table:3}
\end{table}
\setlength{\tabcolsep}{20pt}

\section{Conclusion}
The study has evaluated a 6-channel approach to address the domain-shift issue by incorporating the generated fake images using image-to-image translation. However, we have not achieved the expected results. One possible reason is the quality of the generated images is inferior compared to the original images, especially the fake day images generated from the data of night scenes, as illustrated in Fig. \ref{fig:2} and Fig. \ref{fig:3}. If we merely concatenate the original high-quality images with their inferior counterparts, the model may treat the low-quality fake image channels as some kind of "noise", and thus, the model could hardly learn more useful information from the concatenated 6-channel representations. Another possible reason is that the domain shift issue may still exist in the combined 6-channel representations, which prevents the model from extracting useful information from the concatenated representations. Moreover, the dataset we used in the experiments only has limited samples, which are insufficient to train the model. We hope the idea of augmented data representation can inspire more further investigations and applications.
%
%
%
\bibliographystyle{splncs04}
\bibliography{Myreference}

\end{document}